\documentclass[conference]{IEEEtran}
\IEEEoverridecommandlockouts
\usepackage{cite}
\usepackage{amsmath,amssymb,amsfonts}
\usepackage{tikz}
\usetikzlibrary{shapes.geometric}
\usetikzlibrary{arrows.meta,arrows}
\newcommand{\tabfigure}[2]{\raisebox{-0.85\height}{\includegraphics[#1]{#2}}}
\usepackage{hyperref}
\usepackage{booktabs}
\usepackage{algorithmic}
\usepackage{graphicx}
\usepackage{textcomp}
\usepackage{xcolor}

\def\BibTeX{{\rm B\kern-.05em{\sc i\kern-.025em b}\kern-.08em
    T\kern-.1667em\lower.7ex\hbox{E}\kern-.125emX}}
\begin{document}

\title{Book Cover Synthesis from the Summary\\
}

\makeatletter
\newcommand{\linebreakand}{%
  \end{@IEEEauthorhalign}
  \hfill\mbox{}\par
  \mbox{}\hfill\begin{@IEEEauthorhalign}
}
\makeatother

\author
{\IEEEauthorblockN{Emdadul Haque}
\IEEEauthorblockA{
\textit{Ahsanullah University of Science}\\
\textit{and Technology, Bangladesh} \\
rafathaque1997@gmail.com}
\and
\IEEEauthorblockN{Md. Faraz Kabir Khan}
\IEEEauthorblockA{
\textit{Ahsanullah University of Science}\\
\textit{and Technology, Bangladesh}\\
farazkabir@gmail.com}
\and
\IEEEauthorblockN{Mohammad Imrul Jubair}
\IEEEauthorblockA{
\textit{University of Colorado Boulder,}\\
\textit{United States}\\
mohammad.jubair@colorado.edu}
\and
\linebreakand 
\IEEEauthorblockN{Jarin Anjum}
\IEEEauthorblockA{
\textit{Ahsanullah University of Science}\\
\textit{and Technology, Bangladesh} \\
jarinanjum375@gmail.com}
\and
\IEEEauthorblockN{Abrar Zahir Niloy}
\IEEEauthorblockA{
\textit{Ahsanullah University of Science}\\
\textit{and Technology, Bangladesh} \\
abrar.zahir.niloy100@gmail.com}
}

\maketitle

\begin{abstract}
The cover is the face of a book and is a point of attraction for the readers. Designing book covers is an essential task in the publishing industry. One of the main challenges in creating a book cover is representing the theme of the book's content in a single image. In this research, we explore ways to produce a book cover using artificial intelligence based on the fact that there exists a relationship between the summary of the book and its cover. Our key motivation is the application of text-to-image synthesis methods to generate images from given text or captions. We explore several existing text-to-image conversion techniques for this purpose and propose an approach to exploit these frameworks for producing book covers from provided summaries. We construct a dataset of English books that contains a large number of samples of summaries of existing books and their cover images. In this paper, we describe our approach to collecting, organizing, and pre-processing the dataset to use it for training models. We apply different text-to-image synthesis techniques to generate book covers from the summary and exhibit the results in this paper.
\end{abstract}


\section{Introduction}
A book cover makes an initial impression on potential readers. It serves as a salesman to its audience. The cover must be recognized as a valuable instrument, not just for its original role of protecting and binding the pages of a book but also for its capability to attract readers. An appealing book cover requires a lot of creativity and time to create. Automating the process of producing a book cover can save significant time and effort. Even if not entirely automated, the process of making a book cover will be accelerated if pertinent design inspiration or ideas can be generated. Additionally, it may lower production costs for book covers while improving visual quality. In our work, we aim to produce book cover images with the aim of improving the cover designing process, leveraging modern text-to-image conversion methods. 

Generative Adversarial Networks(GANs)\cite{N_GAN_NIPS2014_5ca3e9b1} is one of the newest machine learning frameworks. Text-to-image synthesis is one of its applications that uses text to predict image data. We wish to generate a book cover from a summary using these text-to-photo synthesis approaches. We also generate book covers using transformers\cite{transformer}. All these methodologies require a dataset with both image and text as a pair; in our case, the cover image and book summary. Currently, there is no dataset available fulfilling this requirement. So, we first create a dataset and apply these techniques. Finally, we compare the findings of the experiments and evaluate them using certain well-known strategies.

In short, our contribution can be noted as follows-- 
\begin{itemize}
\item We present a systematic approach to the creation of a dataset on books. We construct a dataset of English books with $24,786$ unique data entities covering relevant information. Moreover, we present a way of gathering multiple captions for book cover images.
\item We make this dataset publicly available to the research community for further exploration.
\item We train a style-based generator model to produce random book covers. Then, we train some state-of-the-art text-to-image synthesis models to design book covers according to  the synopsis.
\item Lastly, we compare all the outputs and conduct evaluations to conclude.
\end{itemize}

\subsection{Paper Organization}
The paper is organized into the following sections. We review related literature and present our study in Section~\ref{sec:related works}. Our key contributions are shown in Section~\ref{sec:methodology} with the description of our dataset and implementation of different text-to-image methods. Section~\ref{sec:Results} exhibits our results and discusses the evaluations. Finally, we conclude the paper in Section~\ref{sec:conclusion}, and explain our limitations and future research opportunities.

\section{Related Works}
\label{sec:related works}
Image generation from the textual description is an active area of research. Uses of Generative Adversarial Networks\cite{N_GAN_NIPS2014_5ca3e9b1} and transformers\cite{transformer} are observed to accomplish this objective. First, Reed et al. \cite{BasicT2I_reed2016generative} utilized conditional GANs to generate images from text. StackGAN\cite{stylegan} produces imagery from the text by stacking multiple generators and discriminators. The crossmodal attention mechanism is introduced in AttnGAN \cite{Xu_2018_CVPR}  to enable the generator to synthesize pictures with finer text details. Most of the text-to-image GANs used multiple generators and discriminators in their architecture initially. DF-GAN\cite{DF_GAN_2008} introduced a single-stage backbone for text-to-image synthesis. Later, Aditya et al. \cite{DALLE_ramesh2021zeroshot} proposed adopting transformers\cite{transformer} to model image and text tokens as a single data stream to approach this task.

Even though there has been a significant amount of research on text to picture conversion, most of that has been done on the CUB\cite{CUB_Dataset} and COCO\cite{COCO_Dataset} datasets. Images of birds can be found in the CUB\cite{CUB_Dataset} dataset, whereas there are 80 categories in the COCO\cite{COCO_Dataset} dataset. As there is no existing dataset with both images of book covers and relatable captions, creating a book cover from a summary or other description has been an unexplored area. Again, automatic book cover generation, in any form, is a less explored domain. The website Booksby.ai\cite{booksby.ai} uses progressive GANs to create book covers. Wensheng Zhang, Yan Zheng, Taiga Miyazono, Seiichi Uchida, and Brian Kenji Iwana\cite{Towards_Book_Cover_Design_via_Layout_Graphs} proposed developing user-designed book cover graphics in their paper, which combined a layout graph-based generator with SRNet. None of these methods, however, resulted in producing flawless book covers. 


\section{Methodology}
\label{sec:methodology}
We begin by gathering metadata and book cover images by scraping a website. To create a dataset, we clean and organize these data. Our dataset is prepared to train the text to image synthesis models after completing a number of processes. When training some text-to-image GANs, we produce multiple captions from the summary by preprocessing the dataset. We generate book cover designs using these text-to-image conversion methods from the synopsis. Furthermore, a progressive generator model is trained to randomly create cover images from which we can assess the quality of the produced images without relevance to the text.





\subsection{Data Engineering}
We need to gather data and do a number of procedures on it to prepare it for training the text-to-image synthesis models.
\subsubsection{Dataset Preparation} To conduct the experiments, we need a dataset with a book summary and cover image. We choose the Goodreads \cite{goodreads} website because it contains images of book covers, book summaries, titles, and other book-related information. We scrape book metadata and cover images from this website to create our dataset. We scrape and wrangle data in a systematic manner that can be reproduced in order to increase our collection. We construct this dataset in such a way that it can be further used for other tasks and experiments like genre classification, book title generation, and so on. The field name and description of the data we collect are listed in Table~\ref{tab:Dataset Fields}. There are two stages for the preparation process of our dataset: Scraping and Exporting.

\begin{table}[bp]

 \caption{Description of all the field names in the spreadsheet file}
  \centering
  \begin{tabular}{ll}
    \toprule
    \textbf{Field Name} & \textbf{Description}  \\
    \midrule
    title & Title of the book  \\
    author\textunderscore name &  Author of the book \\
    publisher & Publisher of the book \\
    published\textunderscore at & Time of publication  \\
    in\textunderscore language & The language of the book's text  \\
    full\textunderscore summary &  Summary of the entire book \\
    
    image\textunderscore url & URL of the cover image source \\
    source\textunderscore url & URL of the book information source \\
    genres & Genre of the book \\
    genres\textunderscore tags & Other related genre tags \\
    collected\textunderscore from & Website URL of collected data\\
    date\textunderscore time & Time of the data collection \\
    \bottomrule
  \end{tabular}
  
  \label{tab:Dataset Fields}
\end{table}

\paragraph{Scraping Process} The Scraping process is divided into four sub-process. They are--
\begin{itemize}
\item[-] \textit{Script Creation for Automation.} We first analyze the website's markup and create a script according to that to scrape the website to collect essential information and make the data collection process automated.

\item[-] \textit{Raw Data Scraping.} We extract data directly from websites in this procedure. We omit any entity that has corrupted or missing values. We eliminate special characters and apply a format to the date-related field data. We save the collected information into a file.

\item[-] \textit{Data Grouping.} We merge data from the multiple files generated from the previous steps. We organize data according to their genre. We remove duplicates if any exist. Then we save all data in separate files genre-wise.

\item[-] \textit{Final Data Merging.} In this process, we take data from all the files having grouped data. We merge these files and remove duplicates. Finally, we save them in a spreadsheet file.
\end{itemize}

After all of these steps are completed, we are left with a single spreadsheet with all of the relevant data. To ensure that all genres have the same quantity of data, we remove data from different genres at random, balancing them with the genre with the least amount of data. Additionally, we finalize the data wrangling and maintain the data with books that are solely written in English. 

\paragraph{Exporting Process}
In this step, we export data from the spreadsheet file. We download the images from the source URL and store them in a folder according to the genre. We also extract the texts from the spreadsheet file, save them into a text file, and store them in a folder according to the genre. All the text and image file names follow the same naming conventions. The dataset is then divided into two parts: images and text. The images directory contains genre sub-directories with image files, and the text directory contains genre sub-directories with text files. Finally, we save our dataset in a compressed format. 

The whole dataset preparation process is shown in Fig.~\ref{fig:Dataset Preparation}. The dataset can be accessed from this http URL---\href{https://cutt.ly/book-cover-dataset}{cutt.ly/book-cover-dataset}




\subsubsection{Data Statistics}
Our dataset contains $24,786$ unique data objects from six genres, with an equal number of data for each genre. The genres are - Children, Mystery and Thriller, Non-fiction, Romance, Science Fiction, and Young Adult. Each data tuple in the spreadsheet file has 12 attributes. A comparison between CUB\cite{CUB_Dataset}, COCO\cite{COCO_Dataset} and our dataset is shown in Table~\ref{tab:data comparison}. Some sample images from these genres are shown in  Table~\ref{tab:Dataset samples}. 

\begin{table}[bp!]
\caption{Comparison between datasets}
  \centering
  \begin{tabular}{lcc}
    \toprule
    \textbf{Method} & \textbf{Number of Samples }  & \textbf{Class} \\
    \midrule
  
    CUB\cite{CUB_Dataset} & 11,788 & 200 \\
    COCO\cite{COCO_Dataset} & 320,000 & 80 \\
    \midrule
    Our Dataset & 24,786 & 6 \\
    \bottomrule
  \end{tabular}
  
  \label{tab:data comparison}
\end{table}

\begin{table}[!bp]
\caption{Sample images from all the genres in our dataset}
  \centering
  \begin{tabular}{ccc}
    \toprule
    \textbf{Children} & \textbf{Mystery and Thriller} & \textbf{Non-fiction}  \\
    \midrule
    \tabfigure{width=20mm,height=25mm}{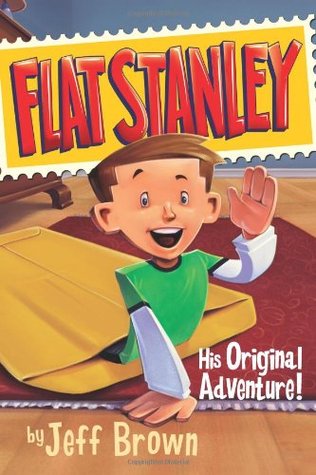}  & \tabfigure{width=20mm,height=25mm}{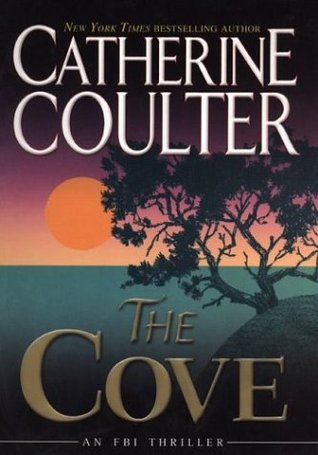} & \tabfigure{width=20mm,height=25mm}{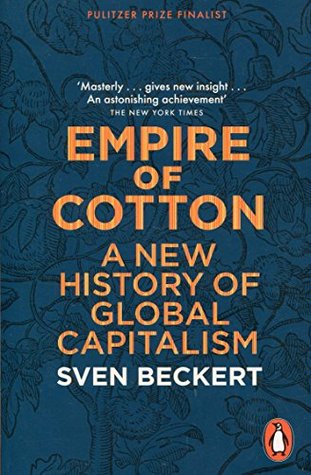} \\
    \midrule
    \textbf{Romance} & \textbf{Science Fiction} & \textbf{Young Adult}  \\
    \midrule
    \tabfigure{width=20mm,height=25mm}{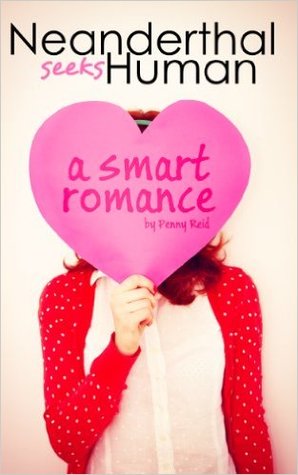}  & \tabfigure{width=20mm,height=25mm}{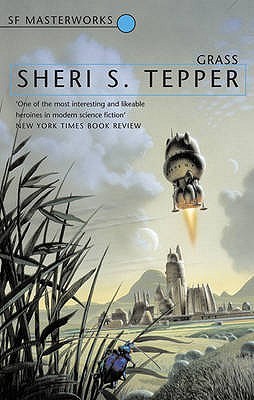} & \tabfigure{width=20mm,height=25mm}{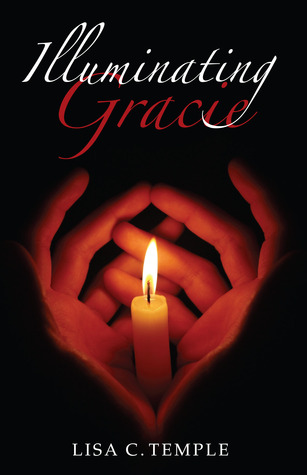} \\
    \bottomrule
  \end{tabular}
  \label{tab:Dataset samples}
\end{table}


\tikzstyle{decision} = [diamond, draw, fill=orange!80,   
    text width=5.0em, text centered, node distance=3.1cm,]   
\tikzstyle{document} = [rectangle, draw, fill=blue!30,   
    text width=4.7em, text centered, node distance=2.5cm,]   
 \tikzstyle{block1} = [rectangle, draw, fill=green!50,  
    text width=8em, text centered, rounded corners, node distance= 2cm, minimum height=3em] 
\tikzstyle{block} = [rectangle, draw, fill=yellow!50,   
    text width=8em, text centered, rounded corners, node distance=2cm, minimum height=4em] 
    
\tikzstyle{blankblock} = [rectangle, draw, fill=yellow!50,opacity=0] 
    
\tikzstyle{line} = [draw, -latex']  
\tikzstyle{cloud} = [draw, ellipse,text width= 2.9em, fill=red!50, node distance=2cm, minimum height=3em]  
  
\tikzstyle{ioi} = [trapezium, draw, trapezium right angle=120,rounded corners, fill=blue!60, node distance=2.8cm, minimum height=2.7em]  
 \tikzstyle{io} = [trapezium, draw, trapezium right angle=110,rounded corners, fill=red!20, node distance=1.9cm, minimum height=2.9em]   
 
\begin{figure}
\centering
\begin{tikzpicture}[node distance = 1.8cm, auto] 
    \node [block] (A) {Source Website};  
    \node [blankblock,below of=A] (hidden1) {};  
    \node [block1, left of = hidden1](B){Collect image URL}; 
    \node [block1, right of = hidden1](C){Collect text data};
    \node [block,below of=hidden1] (D) {Store into a file};
    \node [blankblock,below of=D] (hidden2) {};
    \node [block1, right of = hidden2](E){Download image}; 
    \node [block1, left of = hidden2](F){Text pre-processing};
    \node [block,below of=hidden2] (G) {Collect image and text data and store into separate folders};
    \node [blankblock,below of=G] (hidden3) {};
    \node [block1, left of = hidden3](H){Split data according to the genre}; 
    \node [block1, right of = hidden3](I){Split data without the genre};
    \node [block1, below of = H](J){Create a genre based compressed file}; 
    \node [block1, below of = I](K){Create a compressed file without genre};
    
    \path [line] (A.west) -| (B.north); 
    \path [line] (A.east) -| (C.north);
    \path [line] (B) -| (D);
    \path [line] (C) -| (D);
    \path [line] (D.west) -| (F.north); 
    \path [line] (D.east) -| (E.north);
    \path [line] (E) -| (G);
    \path [line] (F) -| (G);
    \path [line] (G.west) -| (H.north); 
    \path [line] (G.east) -| (I.north);
    \path [line] (I) -- (K); 
    \path [line] (H) -- (J);   
    
\end{tikzpicture}  
\caption{The complete workflow of the dataset preparation process.} 
\label{fig:Dataset Preparation}
\end{figure}



\subsubsection{Data Pre-processing for Text-to-Image GANs}
The datasets previously used to train conventional text-to-image GANs contain multiple captions per image. To have similarity with this, we process the texts of our dataset to ensure multiple meaningful captions per book cover photo. For that, we choose the summary from the primary text file. With BERT \cite{BERT} and T5 \cite{T5} pre-trained models, we generate two abstractive summaries from the raw summary. We also develop an extractive summary using a python library\cite{sumy}. Some description contains information regarding the book's earlier events or versions, which may not be pertinent to the synopsis. So, using these approaches allows the summaries to be condensed while retaining the significance of the terms and lines. Additionally, employing different strategies enables us to produce distinct captions from the same overview. We also consider the title as one of the captions. The other texts are removed, and finally, the text files in the folder contain only these four captions. We do not consider any summary containing characters less than 40 in the primary text file and remove such text-image pairs. Keeping words to a set length maintains the uniformity of the training process.

 

\subsection{Book Cover Generation}

To observe the capabilities of GANs when producing book cover images, we first train a model with our dataset that generates images without being conditioned on text. StyleGAN(V2)\cite{styleganv2} has shown remarkable results when generating random images. We, therefore, decide to use this architecture to train with our dataset and, in this instance, evaluate the outcome. This not only enables us to compare the results of text-to-image synthesis techniques but also allows us to use the outcomes as creative inspiration for cover designs. So, using our dataset, we train StyleGAN(V2)\cite{styleganv2} model to generate book cover pictures with a $128\times128$ pixel resolution. During training, we keep the default training hyperparameters. Some sample results are shown in Table~\ref{tab:Sample images}.\\

\subsubsection{Cover Generation Using Text-to-image GANs}
Different text-to-image GANs have different approaches when generating an image from texts. For our experiments, we choose AttnGAN\cite{Xu_2018_CVPR} and DF-GAN\cite{DF_GAN_2008}, two notable text-to-image GAN architectures. The  AttnGAN\cite{Xu_2018_CVPR} uses a crossmodal attention mechanism while DF-GAN\cite{DF_GAN_2008} is a single-stage backbone for text-to-image synthesis, which means it uses a single generator and a single discriminator.

First, to train AttnGAN\cite{Xu_2018_CVPR} models, we need to train the text encoder on our dataset. We choose three genres of data from our dataset to train AttnGAN\cite{Xu_2018_CVPR}. The genres are - Children, Romance, and Nonfiction. We take $80\%$ of the data to train and rest for testing. After that, we begin by using the approach mentioned above to preprocess the dataset and collect four captions per image. We train the DAMSM\cite{DF_GAN_2008} model as our text encoder with text embedding dimension $64$. Then we use this encoder for our training AttnGAN\cite{Xu_2018_CVPR} model. The DAMSM\cite{DF_GAN_2008} encoder produces word vectors that are visually discriminative and does not experience problems such as clustering word vectors of various colors in the same vector space, which is present in the traditional text-encoders.

Next, we again train the DAMSM\cite{DF_GAN_2008} encoder with all genres of our dataset after collecting four captions per image using our preprocessing method. The text embedding dimension for training the DF-GAN\cite{DF_GAN_2008} model is $256$. We train DF-GAN\cite{DF_GAN_2008} model with similar approaches as we trained AttnGAN\cite{Xu_2018_CVPR} with all genres in the dataset. 

We mostly follow the training process as described in the official publication. The output image resolution for both models was $256\times256$.

\begin{table}[!bp]
\caption{Random book cover samples generated using StyleGAN(V2)\cite{styleganv2}}
  \centering
  \begin{tabular}{ccc}
    \toprule
    \tabfigure{width=25mm,height=25mm}{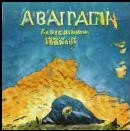}  & \tabfigure{width=25mm,height=25mm}{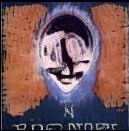} & \tabfigure{width=25mm,height=25mm}{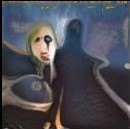} \\
    \midrule
    \tabfigure{width=25mm,height=25mm}{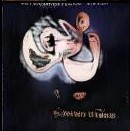}  & \tabfigure{width=25mm,height=25mm}{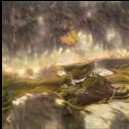} & \tabfigure{width=25mm,height=25mm}{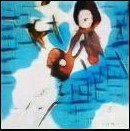} \\
    \midrule
    \tabfigure{width=25mm,height=25mm}{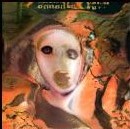}  & \tabfigure{width=25mm,height=25mm}{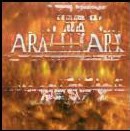} & \tabfigure{width=25mm,height=25mm}{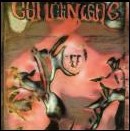} \\
    \bottomrule
  \end{tabular}
  \label{tab:Sample images}
\end{table}
\subsubsection{Cover Generation Using Transformers}
We experiment with a transformer-based approach to produce a cover design from the summary. DALL-E\cite{DALLE_ramesh2021zeroshot}, a transformer-based architecture, has shown impressive results in producing an image from captions. We choose to apply this approach using our dataset to generate book cover images. We train the models with a different strategy this time. We do not preprocess the text in this instance. Instead, we only use the raw summary from the texts and one caption per image. We train this model with all the genres of our dataset. The cover visuals generated from this method are also of size $256\times256$.
Samples generated from the text to image synthesis methods are shown in Table~\ref{tab:Sample t2i images}.

\begin{table}[bp!]
\caption{Images generated from the summary using AttnGAN\cite{Xu_2018_CVPR}, DF-GAN\cite{DF_GAN_2008}, DALL-E\cite{DALLE_ramesh2021zeroshot}}
\centering
  \begin{tabular}{p{0.35\linewidth}  p{0.55\linewidth}}
    \toprule
  
    \multicolumn{1}{c}{\textbf{AttnGAN}} &
    \multicolumn{1}{c}{\textbf{}}  
    \\
    \tabfigure{width=1\linewidth}{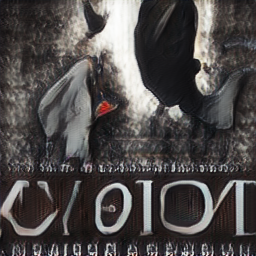}
    &
    \vspace{-5mm}
    Hidden deep in the snowy mountains lives a brotherhood of immortal guardians banished from the heavens. After centuries of exile head guardian raven can taste hell. One wrong move could destroy his soul. When fate assigns him a female mortal to protect his forbidden desire becomes a distraction he ca not afford. Giving in to temptation risks plummeting into the fiery pits of hell dragging her along with him...

  \\

    \midrule
    \multicolumn{1}{c}{\textbf{DF-GAN}} &
    \multicolumn{1}{c}{\textbf{}}  
    \\
    \tabfigure{width=1\linewidth}{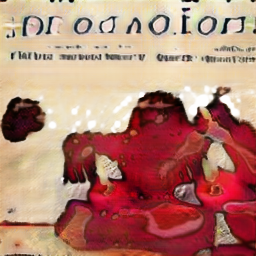}
    &
    \vspace{-5mm}
   The end of exilelong before the world of the ganymeans blew apart millennia ago the strange race of giants had vanished. No one could discover their fate nor where they had gone nor why. There was only a wrecked ship abandoned on a frozen satellite of jupiter. And now earth code and scientists were there determined to ferret out the secret of the lost suddenly spinning out of the vastness of space and immensity of time the ship of the strange humanoid giants returned...

  \\
  \midrule
    \multicolumn{1}{c}{\textbf{DALL-E}} &
    \multicolumn{1}{c}{\textbf{}}  
    \\
    \tabfigure{width=1\linewidth}{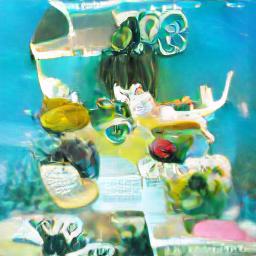}
    &
    \vspace{-5mm}
   And tuppence beresford were restless for adventure so when they were asked to take over blunt international detective agency they leapt at the chance. After their triumphant recovery of a pink pearl intriguing cases kept on coming their way a stabbing on sunningdale golf course cryptic messages in the personal columns of newspapers and even a box of poisoned chocolates.

  \\
    \bottomrule
  \end{tabular}
  \label{tab:Sample t2i images}
\end{table}

\section{Experiments}
\label{sec:Results}
We mostly follow the training process as described in the primary publications. We analyze the outcome qualitatively and quantitatively when the training stage of our experiments is done. Results are presented in Table~\ref{tab:comparison results} and Table~\ref{tab:performance}. 

\begin{table*}[t]
  \caption{\textbf{Book cover designs generated from the summary using AttnGAN\cite{Xu_2018_CVPR}, DF-GAN\cite{DF_GAN_2008}, DALL-E\cite{DALLE_ramesh2021zeroshot} models}}
    \begin{tabular}{cccc}
     \hline 
\multicolumn{1}{c}{

\textbf{Summary}

} &
\multicolumn{3}{c}{\textbf{Generated Image}} 
\\
\hline 
\multicolumn{1}{c}{\textbf{}}  &
\multicolumn{1}{c}{\textbf{AttnGAN}}  &
\multicolumn{1}{c}{\textbf{DF-GAN}}  &
\multicolumn{1}{c}{\textbf{DALL-E}} 
\\

    \multicolumn{1}{p{8.2cm}}{
    Debbie allen contemporary retelling of the classic tale the twelve dancing princesses with illustrations from kadir nelson reverend knight ca not understand why his twelve sons sneakers are torn to threads each and every morning and the boys are not talking. They know their dancing would not fit with their father image in the community. Maybe sunday a pretty new nanny with a knack for getting to the bottom of household mysteries can crack the case...
} & 
          \tabfigure{width=26.5mm,height=26.5mm}{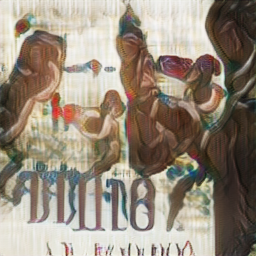}
         & 
           \tabfigure{width=26.5mm,height=26.5mm}{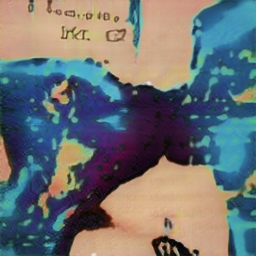} 
          & 
           \tabfigure{width=26.5mm,height=26.5mm}{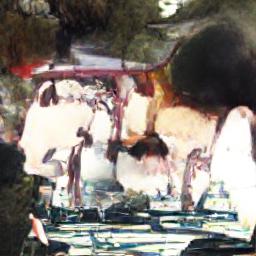} 
          \\
          \addlinespace \hline \addlinespace
         
         \multicolumn{1}{p{8.2cm}}{
   In this explosive account of wrongful acts and jesse ventura takes a systematic look at the wide gap between what the american government knows and what it reveals to the american people. For too long we the people have sat by and let politicians and bureaucrats from both parties obfuscate and lie. And according to this former navy seal former pro wrestler and former minnesota governor the media is complicit in these acts of deception. For too long the mainstream press has refused to consider alternate possibilities and to ask the tough questions...
} & 
          \tabfigure{width=26.5mm,height=26.5mm}{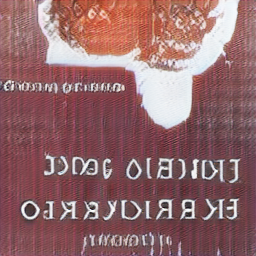}
         & 
           \tabfigure{width=26.5mm,height=26.5mm}{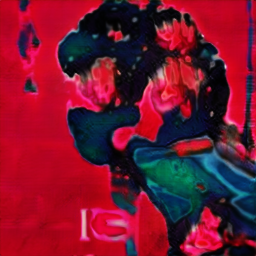} 
          & 
           \tabfigure{width=26.5mm,height=26.5mm}{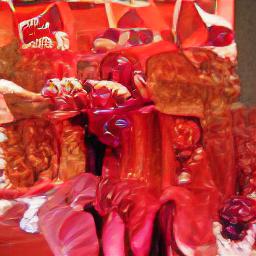} 
          \\
         
         \addlinespace \hline \addlinespace
         
         \multicolumn{1}{p{8.2cm}}{
   A boy who accidentally bonds with a magical beast must set off on an adventure in the mysterious last thing barclay thorne ever wanted was an as an apprentice to the town mushroom farmer barclay need only work hard and follow the rules to one day become the head mushroom farmer himself. No danger required. But then barclay accidentally breaks his town most sacred rule never ever ever stray into the woods for within the woods lurk vicious magical barclay horror he faces a fate far worse than being eaten he unwittingly bonds with a beast and is run out of town by an angry mob... 
} & 
          \tabfigure{width=26.5mm,height=26.5mm}{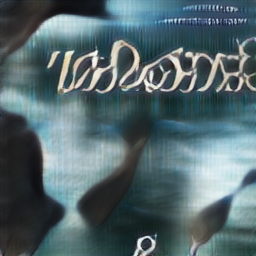}
         & 
           \tabfigure{width=26.5mm,height=26.5mm}{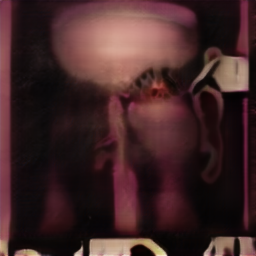} 
          & 
           \tabfigure{width=26.5mm,height=26.5mm}{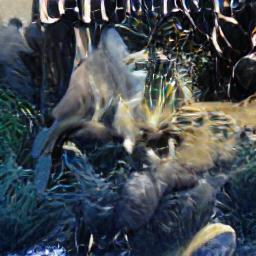} 
          \\

         \bottomrule
          
    \end{tabular}
    \label{tab:comparison results}
\end{table*}

\subsection{Quantitative Evaluation}
For quantitative evaluation of our outputs, we choose Inception Score (IS)\cite{Inception_Score} and Frechet Inception Distance (FID)\cite{Fid_score} following previous works.

\begin{itemize}
 \item \textit{Inception Score(IS).}  The Inception v3\cite{InceptionV3} is a popular deep convolutional image classifier. It correlates well with the human evaluation of the quality of the image. The Inception Score\cite{Inception_Score} computes the  Kullback-Leibler (KL) divergence between the conditional and marginal class distributions using this pre-trained Inception v3 network. Higher IS\cite{Inception_Score} indicates that the produced pictures are of higher quality, and each image distinctly belongs to a specific class.

 \item \textit{Frechet Inception Distance (FID).} The feature vectors of images are used in FID\cite{Fid_score}. It evaluates the quality of images. Based on the extracted features from a pre-trained Inception v3 network, the FID\cite{Fid_score}  compares the distributions of synthetic and natural images and computes the Frechet distance between them. A lower FID\cite{Fid_score} score indicates the images to be more realistic.
\end{itemize}

From our experiments, we find that StyleGAN(V2)\cite{styleganv2} performs better than others in terms of IS score\cite{Inception_Score}, and AttnGAN\cite{Xu_2018_CVPR} performs better in the case of FID\cite{Fid_score}. The results are shown in Table~\ref{tab:performance}.

\begin{table}[b]
\caption{Performance metrics of the models in our experiments}
  \centering
  \begin{tabular}{lcc}
    \toprule
    \textbf{Method} & \textbf{IS$\uparrow$ }  & \textbf{FID$\downarrow$} \\
    \midrule
    StyleGAN(V2)\cite{styleganv2} & \textbf{4.76} & 76.19  \\
    \midrule
    AttnGAN\cite{Xu_2018_CVPR} & 3.98 & \textbf{48.53} \\
    DF-GAN\cite{DF_GAN_2008} & 4.65 & 64.48 \\
    DALL-E\cite{DALLE_ramesh2021zeroshot} & 3.49 & 135.94 \\
    \bottomrule
  \end{tabular}
  
  \label{tab:performance}
\end{table}

\subsection{Qualitative Evaluation}

We compare the outputs generated by AttnGAN\cite{Xu_2018_CVPR}, DF-GAN\cite{DF_GAN_2008}, and DALL-E\cite{DALLE_ramesh2021zeroshot} models. We develop three different book covers from one summary using these three models and observe the images to understand how different strategies differ in terms of output from similar inputs. Some of these results are shown in Table~\ref{tab:comparison results}. Our primary observations tell us that the designs have the appearance and feel of a book cover but do not appear faultless. We see a tendency to mimic the text on top of the cover in terms of AttnGAN\cite{Xu_2018_CVPR} outputs, while this was less apparent in the other methods. We perceive the results generated by DALL-E\cite{DALLE_ramesh2021zeroshot} to be more vibrant. In terms of text-image semantic consistency, a connection between the theme of these outputs and the summary concept is visible in most cases. For instance, when the description discusses beasts or horror, the generated visuals tend to have a darker mood. 

These cover images, overall, look artistic by nature. As the judgment of artistic creation is subjective, it is hard to pick one method from these as the best. However, using different approaches offers us versatility in design and provides different themes and patterns for the cover. Therefore, designers have the freedom to select a method that yields outcomes that suit their preferences.

\section{Conclusion}
\label{sec:conclusion}
In this paper, we demonstrate our application of some text-to-image synthesis techniques to generate a summary-based book cover design. To accomplish this, we first create a dataset of English books in a systematic manner. We craft the dataset in such a way that there may be further uses for our dataset in other research tasks. Then, we demonstrate how to create several captions for book covers. After that, we train different models and generate book covers in an automated manner. Finally, we assess the findings to determine the capability of existing text-to-image generation models to produce book covers. We observe that the generated images have the appearance of the book cover but suffer from severe abnormalities such as object deformation, incorrect object positioning, a strange mix of foreground and background elements, and a distorted title on the cover image. Even though these photos can't be used as a professional book cover, they can certainly be used as creative inspiration or book cover design suggestions, which will benefit the artists when designing book covers. The resulting book cover designs will be relevant to the book because they are based on summaries. The result from StyleGAN(V2)\cite{styleganv2} indicates that there is room for improvement for both the generative models and dataset and that producing perfect book cover images is a difficult undertaking. The outcomes show that the text-to-image synthesis architectures used in the experiments do not perform as well on our dataset as they did on the CUB\cite{CUB_Dataset} or COCO\cite{COCO_Dataset} datasets. More research is also needed to determine whether these models are accurate in producing visuals that connect to the text's inner meaning rather than its outward meaning, like the book cover and summary, as our findings suggest that they may not be. In the future, we intend to increase the quantity of data in our dataset. As text-to-image synthesis methods are evolving continuously, we also want to experiment with some of the latest techniques on our dataset to produce book covers from the summary.


\begin{thebibliography}{00}
\bibitem{sumy}Simple library and command line utility for extracting summary from HTML pages or plain texts. The package also contains simple evaluation framework for text summaries. Implemented summarization methods are described in the documentation. I also maintain a list of alternative implementations of the summarizers in various programming languages. | sumy.
\end{thebibliography}
\end{document}